\documentclass{article}

\usepackage[final]{corl_2024} 
\usepackage{amsmath} 
\usepackage{graphicx}
\usepackage{booktabs}
\usepackage{array}

\title{Triple Regression for Sim2Real Adaptation in Human-Centered Robot Grasping and Manipulation}

\author{
    Yuanhong Zeng\textsuperscript{1}, Yizhou Zhao\textsuperscript{1}, Ying Nian Wu\textsuperscript{1}\\
    \textsuperscript{1}University of California, Los Angeles,\\
    \texttt{\{yuanhongzeng, yizhouzhao\}@ucla.edu}
}

\begin{document}
\maketitle



\begin{abstract}
    Sim2Real (Simulation to Reality) techniques have
gained prominence in advancing robot grasping and manipulation, particularly in human-centered environments where adaptation to dynamic, unpredictable conditions is essential. This paper introduces the Triple Regression framework, which uses a digital twin in real time to enhance robot learning and interaction. The framework addresses the reality gap in human-robot collaboration by: (1) mitigating projection errors between real and simulated camera perspectives through dual regression models and (2) detecting and compensating for discrepancies in robot control using a third regression model. Experiments in picking up and pouring water from random places, which are critical for collaborative robots in dynamic human environments, demonstrate the effectiveness of our method with only the input of an RGB camera. Our work enhances adaptive, real-time robotic decision making in collaborative human-centered tasks and pushes the boundaries of robot learning in both simulation and real-world scenarios.
\end{abstract}



\section{Introduction}
	
Simulation plays a pivotal role in the development and validation of robotic systems. In human-centered applications, the simulation environment can play a critical role in the design and testing of robots. Creating such a simulation environment often encompasses (1) the creation of a \textit{digital twin} representing the actual manufacturing environment; (2) the planning and execution of various robot decision trajectories within the simulation; (3) a thorough assessment of each trajectory's outcomes; and the deployment of the most effective trajectory in a real-world setting. The key is to improve efficiency by automating the aforementioned stages.

Navigating simulation-to-reality (Sim2Real) applications presents two key challenges, often referred to as the \textit{sim2real gap}. First, it is difficult to build an accurate enough digital twin that captures the real-world scenes in a simulation using sensor data (e.g. RGB images from cameras). Recent advances, such as 6D pose estimation \cite{kehl2017ssd, tian2020robust} and 3D scene reconstruction \cite{denninger20203d, mildenhall2021nerf}, have made significant strides in addressing this challenge. However, both approaches have limitations. The 6D pose estimation method faces difficulties in generalizing to unfamiliar objects, while 3D scene reconstruction \cite{zhao2022opend}, although capable of capturing intricate details in complex environments, often blurs the distinction between interactive and stationary objects, obscuring important object-specific details. The second challenge, which arises from the unreal nature of simulated physics and dynamics, lies in the seamless application of the simulation result in real-world situations~\cite{gong2023arnold}. 

To address the challenges above in Sim2Real and digital twin robotic applications, we introduce the triple regression framework for cameras with undetermined intrinsic properties that include focal length, camera resolution, and camera position. 
\begin{itemize}
    \item The first regression aligns the cameras in the real world with those in the simulated environment.
    \item The second regression estimates the target object's position and subsequently places the approximated object into the simulation space by leveraging image segmentation,
    \item The third regression is applied to compensate for errors originating from the simulated robotic trajectory and dynamics. 
\end{itemize}

\begin{figure}[t]
    \centering
    \includegraphics[width = 0.96\textwidth]{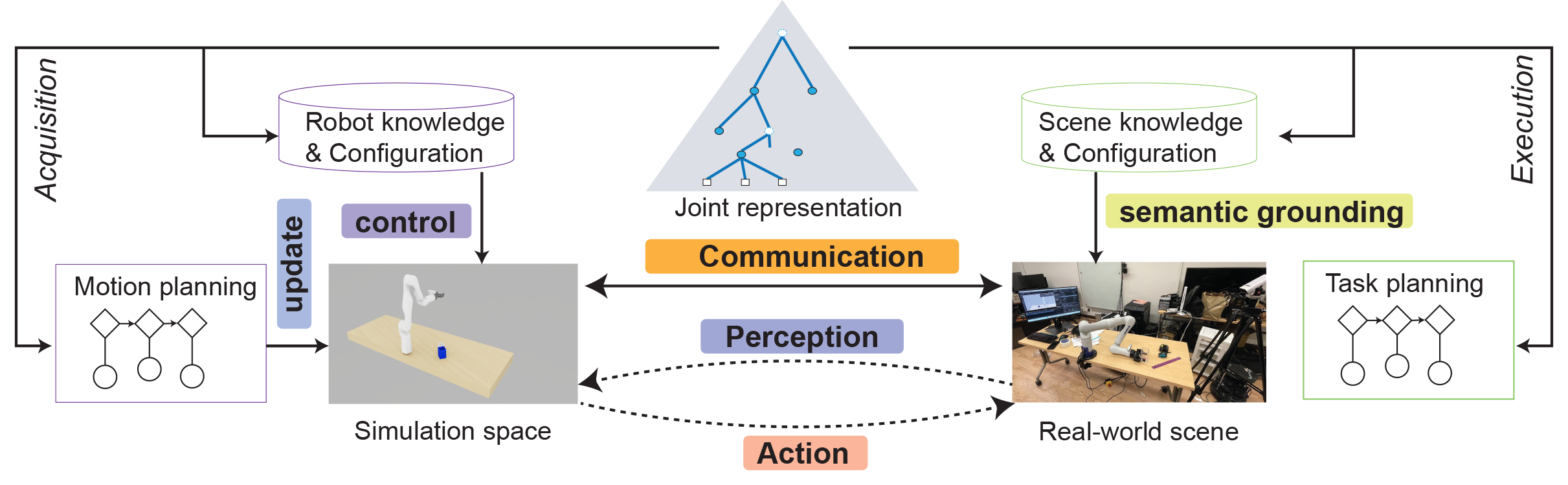}
    \caption{\textbf{Text-to-action framework:} Tasks described language are converted into a joint representation which records spatial, and temporal information of the scene and task. A digital twin is created based on semantic grounding and camera observation. Motions are planned in the simulator. VQA is used to ascertain the success of the performed action}
    \label{fig:teaser}
\end{figure}

Using the triple regression framework, we propose a practical framework for text-to-action tasks as shown in Figure \ref{fig:teaser}. The process begins with using semantic grounding to generate a joint task representation, breaking the task down into multiple grab-and-place actions. Using the segmentation modules \cite{kirillov2023segment}, the dimensions and positions of these 3D objects are determined and integrated into a simulation engine to create a digital twin in real time. Within the simulation, various trajectories are created. A Visual Question Answering (VQA) model ensures task success. Subsequently, successful simulations are implemented in the real world. This methodology offers an automated Sim2Real workflow, notably requiring only a standard RGB camera and basic site measurements.

We conducted an experiment in which we pick up a jar at a random position and pour its water into a cup at another random position. The success of the experiment requires the robot's control of rigid bodies and fluids and the plan to overcome random placement of the camera, robot, jar, and cup. Experimental results show that our methodology provides an offline and zero-shot solution (no training for the robot in real space) and improves the robot's inference speed by at least $50\%$ compared to the state-of-the-art approach. Moreover, our methodology can improve the success rate by $75\%$ under the conditions of randomized object placement.

Our contribution can be summarized as (1) we present a systematic framework to tackle the reality gap stemming from constructing real-time digital twins and migrating simulated trajectory to reality; (2) we propose a Sim2Real solution for the robot manipulation tasks; (3) we demonstrate its effectiveness through physical experiments on picking up and pouring tasks using a Kinova Gen 3 robot.

\section{Method}

\subsection{Joint task planning module}

\begin{figure}[t]
    \centering
    \includegraphics[width = 0.95\linewidth]{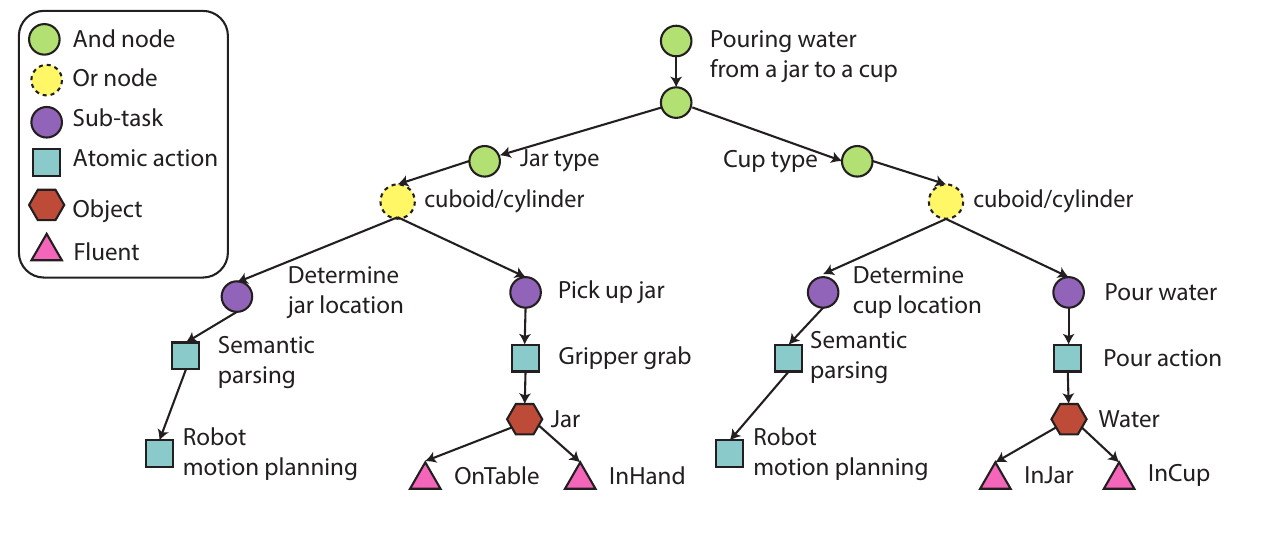}
    \caption{\textbf{Representation of pouring water from a jar to a cup with STC-AOG}. The data structure reduces a sentence into subgoals for task planning. }
    \label{fig:aog}
\end{figure}

The spatial-temporal-causal And-Or graph (STC-AOG)~\cite{xiong2016robot} is used as a task planning model to represent the goal and the current cognitive state of the robot. This structure allows us to associate spatial, temporal, and causal relationships within the task description. The task of \textit{pouring water from a jar from a random position to a cup at another random position}, is shown in Figure \ref{fig:aog}.

In general, an And-Or Graph consists of nodes and edges. The set of nodes includes Or node, And node, and Terminal node. Each \textbf{Or node} specifies the Or relation: only one of its children nodes would be performed at a given time. An \textbf{And node} represents the And relation and is composed of several children nodes. Each \textbf{Terminal node} represents a set of entities that cannot be further decomposed. The edge represents the top-down sampling process from a parent node to its children nodes. The root node of the And-Or tree is always an And node connected to a set of And/Or nodes. Each And-node represents a sub-task which can be further decomposed into a series of sub-tasks or atomic actions.


\noindent \textbf{Causal relation.} 
Causal knowledge represents the pre-conditions and the post-effects of atomic actions. We define this knowledge as fluent changes caused by an action. Fluent can be viewed as some essential properties in a state that can change over time, e.g., the temperature in a room and the status of a heater. For each atomic action, there are pre-conditions characterized by certain fluents of the states. For example, the water in the experiment should initially be in the jar and then transferred to the cup.

\noindent \textbf{Temporal relation.}
Temporal knowledge encodes the schedule for an agent to finish each sub-task. It also contains the temporal relations between atomic actions in a low-level sub-task. The sub-task of preparing in this study, for example, consists of \textit{picking up jar} and \textit{pouring water from the jar to the cup}.

\noindent \textbf{Spatial relation.}
Spatial knowledge represents the physical configuration of the environment that is necessary to finish the task. In our case, to pick up the jar, the robot needs to know the location and the rotation jar. This is determined by the following triple regression framework.

\subsection{Triple Regression for Matching Simulation and Reality}

\begin{figure*}
    \centering
    \includegraphics[width = \textwidth]{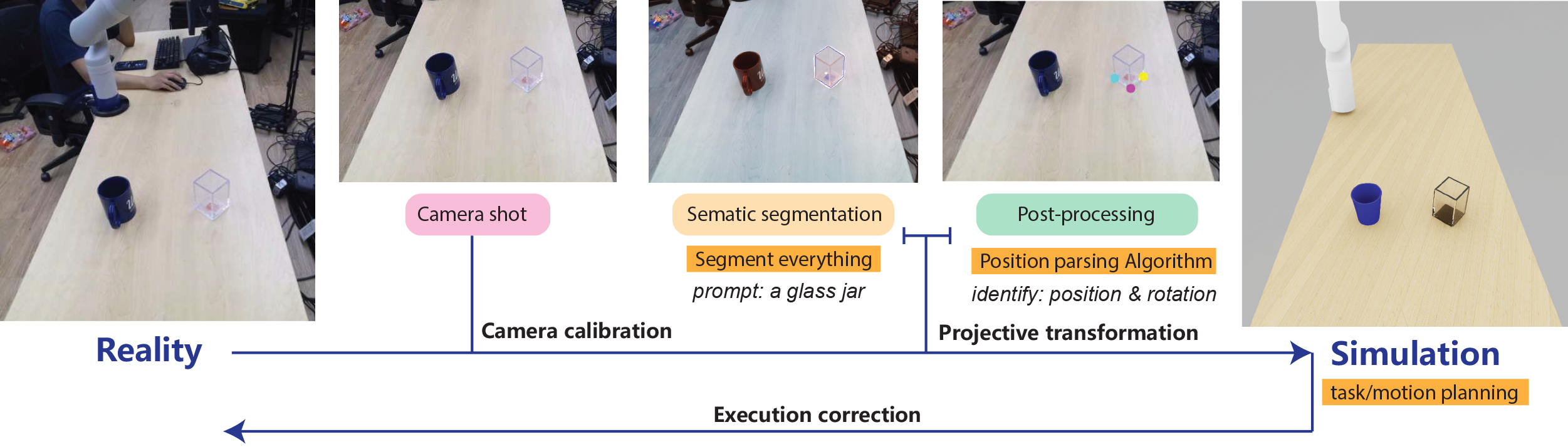}
    \caption{\textbf{Workflow of triple regression framework.} The framework creates digital twins by 1) camera shots, 2) Semantic-based segmentation and contour extraction, 3) key point identification, and 4) object placement in simulation. Plans are generated and simulated. Coordinates of successful plans are corrected and executed in reality.}
    \label{fig:triple_regression}
\end{figure*}

The triple regression framework is designed to align the virtual environment with the real world. The workflow is shown in Fig \ref{fig:triple_regression}. The framework addresses reality gaps that are caused by (1) intrinsic discrepancies between the real and simulated camera, (2) errors while comprehending scenes and determining object locations, and (3) differences in executing planned trajectories between simulated robot and real robot. Equation \ref{eq:matching} gives a formal definition of aligning the parse graph in simulation, $pg^s$, with the parse graph in reality $pg^r$, by aligning the camera, scene, and robot control separately. 

\begin{equation}\label{eq:matching}
    P(pg^s | pg^r) = \prod_{i \in \{\text{camera},\text{scene}, \text{robot}\}} P(pg^s_{i} |  pg^r_{i}) 
\end{equation}

We define our symbols as follows. for a point $\textbf{p}=(x, y, z)$ and its homogenous representation $\Tilde{\textbf{p}} = (x, y, z, 1)$, we define $\textbf{u} = (x_r, y_r)$ to be the point in image captured by the real camera and $\textbf{v} = (x_s, y_s)$ to be the point in image captured by simulated camera. Their representations in homogeneous coordinate is $\Tilde{\textbf{u}} = (x_r, y_r, 1)$ and $\Tilde{\textbf{v}} = (x_s, y_s, 1)$.


\noindent\textbf{Camera alignment.} The goal of the first regression is to align real and simulated cameras. In other words, we aim to find a projection from $\textbf{v}$ to $\textbf{u}$. In a projective camera model~\cite{hartley2003multiple}, this projection can be modeled as in Equation \ref{eq:proj}, where $M$ is defined in Equation \ref{eq:matrix}.

\begin{equation}
\label{eq:proj}
\Tilde{\textbf{u}} = M\Tilde{\textbf{v}}
\end{equation}

\begin{equation}
\label{eq:matrix}
    M = \begin{bmatrix}
        A_{2\times 2} & b_{2\times 1} \\
        c_{1\times 2} & 1 \\
        \end{bmatrix}
\end{equation}

To estimate $M$, we construct a reference scene (table and camera) to mirror reality. 2D locations of multiple reference points (eg. table corner and table center) captured in both simulated and real cameras are recorded. $M$ can then be derived with constrained multivariate regression. 

Besides projective transformation, we also consider the 2D scaling and affine transformation. Table~\ref{tab:regression} provides the final comparative analysis of the three regression methods.

\noindent\textbf{Camera calibration and object allocation} The goal of this phase is to estimate the intrinsics of the simulated camera. Using this we can determine the 3D location of a point with 2D pixel locations. Camera in the simulation environment can be modeled as a matrix $C$ formed by multiplying intrinsic $I$ and extrinsic $E$ defined in equation \ref{eq:intrinsic} and \ref{eq:extrinsic}.

\begin{equation}
\label{eq:intrinsic}
I = \begin{bmatrix}
        f_x & 0 & c_x & 0\\
        0 & f_y & c_y & 0 \\
        0 & 0 & 1 & 0\\
        \end{bmatrix}	
\end{equation}

\begin{equation}
\label{eq:extrinsic}
E = \begin{bmatrix}
    R_{3\times 3} & t \\
    0_{1\times 3} & 1
\end{bmatrix}
\end{equation}

With this model, capturing an image can be modeled as transforming point world position $\Tilde{\textbf{p}}$ to pixel position $\Tilde{\textbf{v}}$ as 

\begin{equation}
\label{eq:perspective}
    \Tilde{\textbf{v}} = IE\Tilde{\textbf{p}} = C\Tilde{\textbf{p}}
\end{equation}

The rotation matrix $R$ and 3D translation $t$ in Equation \ref{eq:extrinsic} can be obtained in the simulation environment. By sampling multiple points in the simulation and camera, $I$ can be estimated by solving constrained least squares problems.

In our setup, all points were placed on planes (eg. table) with known normal $\textbf{n} = (a, b, c)$. Combining this constraint, we can solve for $\Tilde{\textbf{p}}$ given $\Tilde{\textbf{v}}$ in Equation \ref{eq:perspective} to determine the 3D location of a point.

Segmentation modules are used to place the object. Anchor points and bounding boxes are generated with pattern matching or text-to-image grounding models \cite{liu2023grounding}. Contour lines of the objects are extracted using \cite{kirillov2023segment} and key points are subsequently determined for cuboid and cylindrical objects. We use these key points to align objects in simulation and reality. 

\noindent\textbf{Regression to accommodate miscellaneous errors.} Another source of error is control discrepancy and inaccuracy in constructing reference objects in the simulation. We employ another regression model (Equation~\ref{eq:offset_regression}) to correct for robot movements. We collect points $\textbf{p}_s$ and manually adjust their \textit{should-be} position $\textbf{p}_r$ during robot execution. Coefficient $D$ can be obtained by solving regression problems. Note that $\Tilde{\textbf{p}}_s$ and $\Tilde{\textbf{p}}_r$ are the homogeneous representation of $\textbf{p}_s$ and $\textbf{p}_r$

\begin{equation}
    \label{eq:offset_regression}
\Tilde{\textbf{p}}_s = D\Tilde{\textbf{p}}_r 
\end{equation}

\subsection{Validating Sim2Real matching}

Our approach to aligning simulation with reality during robot execution involves a multi-step strategy. 

Firstly, we emphasize the dynamic nature of conditions, termed as \textit{fluents}, which can evolve over time. To capture these evolving states, we continuously track changes in fluent at multiple time points to ensure synchronization between the simulated and real-world environments. 

Secondly, we introduce a structured questionnaire that prompts users to provide binary true-false responses pertaining to the observed fluents. This questionnaire serves as a reliable ground truth reference, enabling us to quantitatively assess the correctness of simulation outputs. The questionnaire consists of four questions. (1) Is the robot ready to pick up the jar? (2) Has the robot already picked up the jar? (3) Is the jar above the cup? (4) Is the robot pouring water into the cup?

\begin{figure*}[t]
    \centering
    \includegraphics[width = 0.98\textwidth]{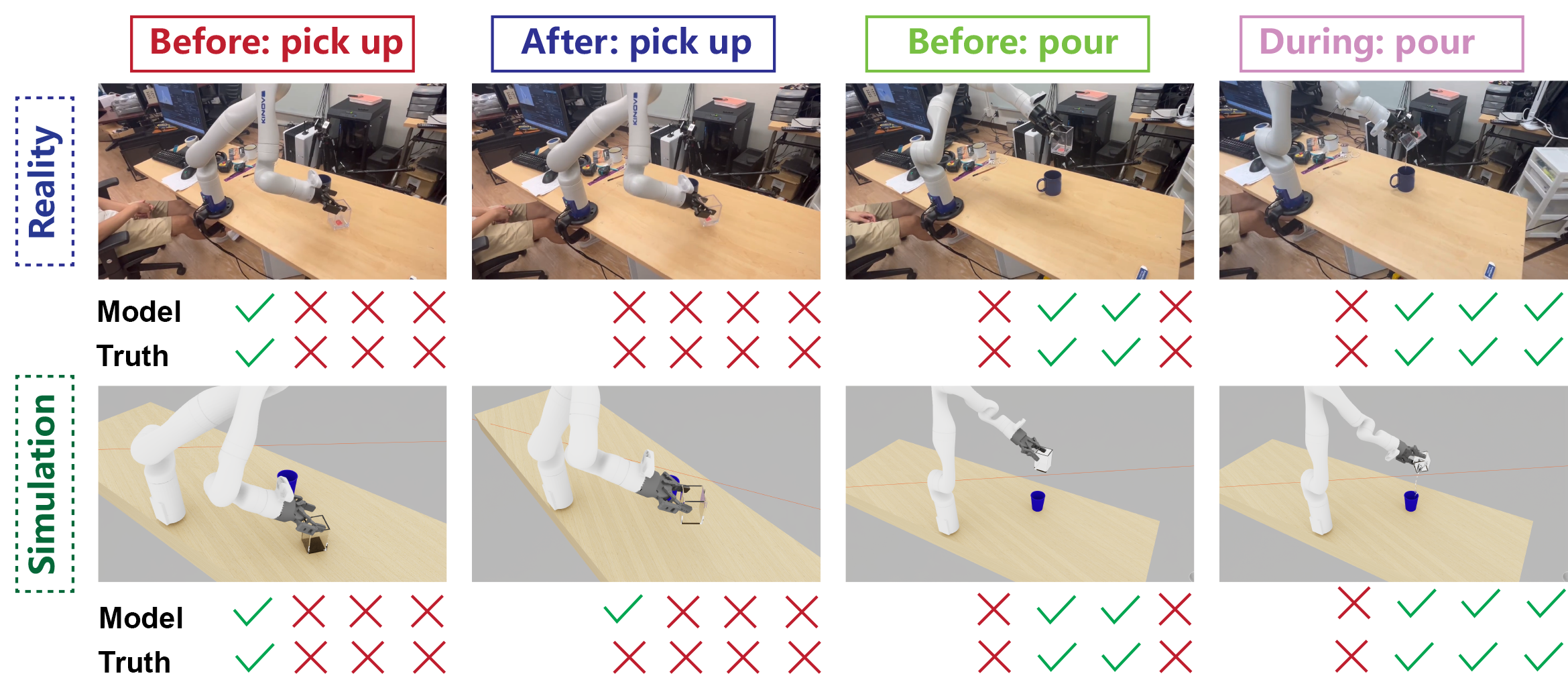}
    \caption{\textbf{Query fluent matching in robot execution}. We select four crucial moments during the execution of the robot in both simulation and reality and report GPT-4V's VQA query answers for the questionnaire: (1) Is the robot ready to pick up the jar? (2) Has the robot already picked up the jar? (3) Is the jar above the cup? (4) Is the robot pouring water to the cup? Output from VQA models and ground truths are shown}
    \label{fig:fluent}
\end{figure*}

Lastly, we leverage advanced Visual Question Answering (VQA) models to analyze and interpret the questionnaire responses. This model takes in an image of each fluent and answers the aforementioned prompts. An example of fluent matching is shown in \ref{fig:fluent}

\begin{equation}\label{eq:fluent}
    P(pg^r | pg^s) = \exp (-\sum_{i} |F^r_i - F^s_i|)
\end{equation}

Given the parse graph in simulation $pg^s$, its discrepancy with the parse graph, in reality, $pg^r$ is estimated quantitatively by the hamming distance between fluents in reality and fluent in simulation as shown in equation \ref{eq:fluent}. $F^r_i$ (fluent in reality) and $F^s_i$ (fluent in simulation) are the VQA query results (true/false) for each question $i$ in the questionnaire above.

\section{Experiment}

The experiment section is organized into three parts: (1) ablation studies on the triple regression framework, (2) grasp-and-pour experiments on a physical robot, and (3) comparisons of VQA models.

\subsection{Ablation study on the triple regression framework}

In this experiment, we validate the necessity of using the projective transformation model perspective discrepancy between real and simulated cameras. A consumer-grade Microsoft RGB camera is mounted in some known location and direction above the table. The table is modeled in Nvidia Omniverse corresponds to the real dimension, and the camera is placed with similar extrinsic that mirror the reality. We align the cameras using three types of regression: full projective transformation, 2D scaling ($A=0$ and $c=0$ in Equation \ref{eq:matrix}), and affine ($c=0$ in Equation \ref{eq:matrix}) A cube with known dimension is placed at known locations both in reality and in simulation. We then compare the pixel location of the 6 corners of the cube captured both in simulation and reality. Figure \ref{tab:regression} summarizes the correlation between the degree of freedom and the achieved Mean Square Error (MSE), signifying the necessity of applying the projective transformation. Using this approach, errors from Sim2Real point position mapping can be controlled as low as 2.7 pixels on average.

\begin{table}[ht]
\centering
\caption{Comparison of regression models used in camera alignment.}
\begin{tabular}{@{}l|cc@{}}
\toprule\toprule
\textbf{Regression method} & \textbf{Degree of Freedom} & \textbf{MSE}  \\ \midrule
Simple linear              & 2                         & $8.2 \pm 4.3$ \\
Affine                     & 6                         & $5.1 \pm 3.2$ \\
Projective                 & 8                         & $2.7 \pm 2.8$ \\ \bottomrule\bottomrule
\end{tabular}

\label{tab:regression}
\end{table}

\subsection{Grasp and pour experiment}

In this section, we first introduce our experiment settings. Subsequently, we explore multiple baseline models for robot planning and control. We then compare our method with the baseline and human performance, and finally, we summarize the results and failure cases.

(1) \textit{Setup}: We conducted pick-and-pour experiments using a Kinova Gen3 7-dof robot, equipped with a Robotiq 2f-85 gripper (jaw length 8.5cm), mounted on a table. The setup included a transparent jar (side length 6.5cm) and a blue cup (radius 8cm). For simulation purposes, we employed the Isaac Sim platform powered by the PhysX engine, where we replicated the table and robotic arm. To align the cameras, we compared their relative position in reality to that in Isaac Sim. Lula implementation of RRT (Rapidly-exploring random tree) \cite{kuffner2000rrt} was used to generate multiple collision-free paths. RMPflow was employed \cite{ratliff2018riemannian} to generate smooth policies to control the robot both in the real world and in simulation. Following this, we input the command "Pour water from the transparent jar into the blue cup" into our Isaac Sim extension and monitored the outcome.

(2) \textit{Baseline models}: We introduce two state-of-the-art models that serve as the robot motion planning module. The pre-trained \textbf{6D-CLIPort}~\cite{zheng2022vlmbench}, which deals with multi-view observations and language input and output a sequence of 6 degrees of freedom (DoF) actions. Another reinforcement learning-based model planning (\textbf{RL})~\cite{mu2021maniskill}, predicts the position and pose of the robot gripper (end-effector) given the vision inputs and the target object name. We also consider the \textit{\textbf{heuristic}} results, i.e. to control the robot motion with the aid of human experts. This human-guided heuristic not only serves as a benchmark of discernment but also offers a benchmark to compare human mastery and our automated Sim2Real process.

(3) \textit{Result}: The analysis of our experiment's results reveals a compelling advancement in both the realms of task execution speed and task success rate.

\begin{figure}[h]
    \centering
    \includegraphics[width = 0.45\textwidth]{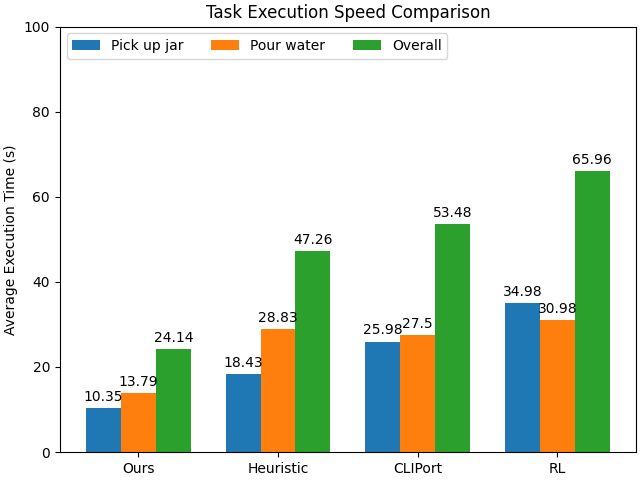}
    \caption{\textbf{Task execution speed comparison.} We compare our method with 6D-CLIPort, RL-based robot motion planning, and human-guided heuristic control, and we measure the average execution time for the whole task.}
    \label{fig:execution_comparison}
\end{figure}

As shown in Figure~\ref{fig:execution_comparison}, our method is remarkably time-saving, reducing the temporal requirements by a significant margin. Specifically, compared by state-of-the-art RL \cite{mu2021maniskill} and 6D-CLIPort models \cite{zheng2022vlmbench}, our methods exhibit a time reduction of $54.5\%$. Even compared with the human-guided heuristic robot control, our methods exhibit a noteworthy $43.4\%$ reduction in time requirements. This substantial efficiency gain proves the efficacy of our approach in robotic decision-making and manipulation.

In terms of task success rate (see Table~\ref{tab:task_success}), the 6D-CLIPort model exhibits a success rate of $25\%$ for both \textit{Pick up} and \textit{Pour water} tasks. Similarly, the RL model showcases slightly improved results, with success rates of $35\%$ for picking up and $40\%$ for pouring water. In contrast, our proposed method showcases a notable leap in performance, achieving a remarkable $75\%$ success rate for picking up and $70\%$ for pouring water. This heightened performance substantiates the robustness and efficacy of our approach in executing these intricate maneuvers.

\begin{table}[t]
\centering
\caption{Comparative analysis of four distinct methods' performance of the pick-and-pour task.}
\begin{tabular}{@{}l|cccc@{}}
\toprule\toprule
\textbf{Method}    & \multicolumn{1}{l}{\textbf{Pick up}} & \multicolumn{1}{l}{\textbf{Pour water}} & \textbf{\begin{tabular}[c]{@{}c@{}}Success \\ (soft)\end{tabular}} & \textbf{\begin{tabular}[c]{@{}c@{}}Success \\ (hard)\end{tabular}} \\ \midrule
\textbf{CLIPort}   & 5/20                                 & 5/20                                    & 5/20                                                                & 4/20                                                               \\
\textbf{RL}        & 7/20                                 & 8/20                                    & 8/20                                                                & 7/20                                                               \\
\textbf{Ours}      & 15/20                                & 14/20                                   & 14/20                                                               & 12/20                                                              \\ \midrule
\textbf{Heuristic} & 19/20                                & 17/20                                   & 17/20                                                               & 16/20                                                              \\ \bottomrule\bottomrule
\end{tabular}
\label{tab:task_success}
\end{table}
\begin{table*}[t]
\centering
\caption{VQA comparison: precision, recall, and F1 score by comparing the model prediction and ground truth}
\setlength{\tabcolsep}{3pt} 
\small 
\begin{tabular}{l|>{\centering\arraybackslash}p{1cm}>{\centering\arraybackslash}p{1cm}>{\centering\arraybackslash}p{1cm}|>{\centering\arraybackslash}p{1cm}>{\centering\arraybackslash}p{1cm}>{\centering\arraybackslash}p{1cm}|>{\centering\arraybackslash}p{1cm}>{\centering\arraybackslash}p{1cm}>{\centering\arraybackslash}p{1cm}|>{\centering\arraybackslash}p{1.5cm}}
\toprule\toprule
         & \multicolumn{3}{c|}{\textbf{Real scene}}                 & \multicolumn{3}{c|}{\textbf{Simulation}}                 & \multicolumn{4}{c}{\textbf{Overall}}                                                                 \\ \cline{2-11} 
         & \textbf{P}$\uparrow$ & \textbf{R}$\uparrow$ & \textbf{F1} & \textbf{P}$\uparrow$ & \textbf{R}$\uparrow$ & \textbf{F1} & \textbf{P}$\uparrow$ & \textbf{R}$\uparrow$ & \multicolumn{1}{c|}{\textbf{F1}} & \textbf{Consistency}  \\ \hline
ViLT     & $50.0\%$           & $46.2\%$        & $48.0\%$          & $42.9\%$           & $54.6\%$        & $48.0\%$          & $46.2\%$           & $50.0\%$        & \multicolumn{1}{c|}{$48.0\%$}          & $87.5\%$              \\
MiniGPT4 & $37.5\%$           & $46.2\%$        & $41.2\%$          & $50.0\%$           & $30.8\%$        & $38.1\%$          & $41.7\%$           & $38.5\%$        & \multicolumn{1}{c|}{$40.0\%$}          & $75.0\%$              \\
GPT-4V    & $\mathbf{85.7\%}$           & $\mathbf{92.3\%}$        & $\mathbf{88.9\%}$          & $\mathbf{85.7\%}$           & $\mathbf{85.7\%}$        & $\mathbf{85.7\%}$          & $\mathbf{85.7\%}$           & $\mathbf{88.9\%}$        & \multicolumn{1}{c|}{$\mathbf{87.3\%}$}          & $\mathbf{96.9\%}$             \\ \bottomrule\bottomrule
\end{tabular}
\label{tab:fluent_query}
\end{table*}

Furthermore, focusing on the \textit{Success (soft)} (allowing a little bit of water leak) and \textit{Success (hard)} (not allowing water leak) metrics, our method again achieves a success rate of $70\%$ and $60\%$, respectively. While the 6DCLIPort and RL models exhibit success rates ranging from $20\%$ to $40\%$ in these categories, our approach consistently demonstrates a superior ability to accomplish the task's objectives, even under more stringent conditions.

However, it's crucial to acknowledge the human-guided heuristic method's exceptional performance, which stands out as a benchmark in this evaluation. With success rates of $95\%$ for \textit{Pick up} and $85\%$ for \textit{Pour water}, as well as $85\%$ and $80\%$ success rates for \textit{Success (soft)} and \textit{Success (hard)}.

(4) \textit{Failure cases}: The major failure case (5 of 8) is when the cup or jar lies beyond the robot's configuration space. In such case, no plan generated inside the simulation environment reaches the goal and thus no actions will be performed in the real world. Another failure case (3 of 8) is when the segmentation module cannot output smooth contours for subsequent determination of key points, causing our key point identification algorithm to generate inaccurate results.

In summation, the experiment results underscore our proposed method's substantial advancements in achieving successful task execution, positioning it as a promising contender against established models. While the human-guided heuristic approach remains the gold standard, our approach showcases remarkable potential.

\subsection{Comparison of VQA models}

We delve into three VQA models to get query results from the questionnaire. The Vision-and-Language Transformer (\textbf{ViLT})~\cite{kim2021vilt} stands at the intersection of visual and linguistic understanding, exhibiting a remarkable capability to seamlessly process and comprehend both visual and textual information. On another model \textbf{MiniGPT-4}~\cite{zhu2023minigpt} emerges as a compact yet potent architecture consisting of the vision transform and large language models, showcasing impressive text generation prowess within a more resource-efficient framework. Meanwhile, GPT-4~\cite{2303.08774}, the latest iteration of the renowned GPT series, continues to push the boundaries of large-scale language modeling. Boasting an extensive knowledge base, GPT-4 demonstrates an unparalleled aptitude for natural language understanding, generation, and manipulation, thus serving as a pivotal milestone in the progression of AI-driven language processing and generation. We apply the visual input API (\textbf{GPT-4V}) of the GPT-4 for inference tasks.

Table~\ref{tab:fluent_query} presents a comprehensive comparison of precision, recall, and F1 scores for query fluent results across real scenes, simulation, and overall performance. Notably, GPT-4V exhibits the highest performance across multiple metrics, boasting an impressive precision, recall, and F1 score in both real scene and simulation scenarios. ViLT and MiniGPT-4 also showcase varying degrees of performance, with GPT-4V demonstrating superior consistency in matching simulation with reality, as evidenced by its remarkable 96.9\% consistency score. These results underline the efficacy of GPT-4V in achieving a harmonious alignment between reality and simulation during robot execution.




\section{Conclusion}

To tackle the enduring challenges of connecting simulation to reality in robotic tasks, we introduce the triple regression framework. This framework is designed to bridge the discrepancies arising from differences in camera parameters, projections, and control dynamics between simulated and real-world settings. Utilizing this framework, we propose an innovative Sim2Real technique for managing robotic grasping and manipulation. The robustness of our approach is demonstrated through our grasp and pour experiments. Additionally, our integration of Vision-Language Models, And-Or graphs for task planning, and vision question and answering opens up new doors for advancing future robotic applications.




\bibliography{example}  
\end{document}